\documentclass{article} 
\usepackage{arxiv_iclr2027_conference,times}

\usepackage{amsmath,amsfonts,bm}

\def\eqref#1{equation~\ref{#1}}
\def\Eqref#1{Equation~\ref{#1}}

\def\1{\bm{1}}

\DeclareMathAlphabet{\mathsfit}{\encodingdefault}{\sfdefault}{m}{sl}
\SetMathAlphabet{\mathsfit}{bold}{\encodingdefault}{\sfdefault}{bx}{n}

\DeclareMathOperator*{\argmin}{arg\,min}

\usepackage{hyperref}
\usepackage{url}
\usepackage{hyphenat}

\title{\nohyphens{Revealing the Biases of Large Language Models by Inferring Prior Distributions}}

\author{Joshua Ryo Cheng$^1$, Thomas L. Griffiths$^{2,3}$ \&  Jake C. Snell$^2$ \\ 
$^1$Department of Mathematics, Princeton University \\
$^2$Department of Computer Science, Princeton University \\
$^3$Department of Psychology, Princeton University\\
\texttt{\{jc4120,tomg,jsnell\}@princeton.edu} \\
}

\usepackage{amssymb}
\usepackage{amsmath}
\usepackage{amsthm}
\theoremstyle{plain}
\newtheorem{theorem}{Theorem}[section]

\newtheorem{corollary}[theorem]{Corollary}

\theoremstyle{definition}
\newtheorem{definition}[theorem]{Definition}

\usepackage{algorithmic}
\usepackage[ruled,vlined]{algorithm2e}
\usepackage{graphicx}
\usepackage{enumitem}

\newcommand{\JS}[1]{}

\iclrfinalcopy 
\begin{document}

\maketitle

\begin{abstract}
Large Language Models (LLMs) have been known to exhibit a wide range of social biases including name-nationality associations, occupation-gender associations and racial stereotypes. Existing approaches to evaluating these biases typically involve measuring surface-level behavior where a model is fed a dataset of prompts and its outputs are scored relative to ground-truth answers. Inspired by prior elicitation in cognitive science, we propose a different perspective that reconstructs latent bias as a prior underlying the model's behavior. We formulate the problem as a Bayesian inverse problem where, given prompt-response observations on a bias benchmark dataset, we infer a low-dimensional latent distribution that summarizes the model's stereotyping tendencies. This paper develops the mathematical foundation, a practical reconstruction algorithm, and experiments using the Bias Benchmark for QA (BBQ) benchmark to infer the prior distributions of LLMs over responses in ambiguous social contexts.
\end{abstract}

\section{Introduction}\label{sec:intro}

LLMs often produce outputs that reflect potentially harmful stereotypes and biases present in training data \citep{kotek2023gender,gallegos2024bias,bai2025explicitly,hu2025generative}. To evaluate social bias in NLP systems, prior work has developed association tests and benchmark datasets designed to elicit stereotyping \citep{nangia2020crows,nadeem2021stereoset,parrish2022bbq}. While these benchmarks provide valuable measurements, they typically summarize behavior through scores computed from generated text or from the observable token probability assignments. Thus, they measure the extent to which a model exhibits certain biases, but fail to explain why the preferences arise.

In this paper, we propose a different perspective. Rather than treating bias benchmarks only as evaluation metrics, we take a Bayesian approach and use them to probe language models in a way that lets us reconstruct a latent space that describes the models' prior beliefs. Concretely, we tackle the question: \textit{Can we reconstruct a latent prior distribution over possible biased responses that explains an LLM's stereotypical preferences?} \JS{Add a sentence on why this matters, foreshadowing BBQ findings.}

A Bayesian framing for this problem is natural because LLMs are inherently probabilistic systems that predict a probability distribution over tokens conditioned on previous ones \citep{bengio2003neural,radford2019language,siddiqui2025functionalapprox}. The motivation for this approach comes from prior elicitation methods in cognitive science, where behavioral judgments have been used to infer the prior distributions that best explain human predictions \citep{garthwaite2005statistical,ohagan2006uncertain}. In this line of work, observed judgments can be used to diagnose which priors and prior parameters people use in certain task domains \citep{griffiths2006optimal}. Related ideas have also been extended to LLMs where prompt-based probing has been used to elicit prior distributions from language models in specific domains \citep{zhu2024eliciting}. 

Building on these ideas, we introduce a Bayesian surrogate model in which an LLM behaves as if it first samples a discrete latent ``bias state'' and then generates outputs according to a state-conditional response model. The central unknown is the prior distribution that summarizes how the model allocates probability across bias hypotheses in social contexts. We interpret this distribution as a summary of the systematic preference structure in an LLM. It is important to note that we are not assuming that LLMs are inherently Bayesian in that they are using Bayesian inference to construct their outputs. Instead, we use a Bayesian surrogate model to analyze and explain the latent prior beliefs that are induced by the model's parameters.

Mathematically, the Bayesian inverse problem framework provides a principled theoretical background for latent prior reconstruction \citep{Dashti2017}. Given observations derived from token log-probabilities on candidate prompt completions, we define a likelihood function for the model and obtain a posterior distribution that approximates the desired latent prior. The main contributions of this paper are:
\begin{itemize}[leftmargin=1em,labelsep=0.25em]
    \item A mathematical formulation for latent prior reconstruction as a Bayesian inverse problem.
    \item A practical reconstruction algorithm based on token log-probabilities using Bayesian inference.
    \item An empirical demonstration using the Bias Benchmark for QA (BBQ) dataset~\citep{parrish2022bbq}.
\end{itemize}

\section{Related Work}
\label{sec:related}

\textbf{Quantifying Social Bias in Language Models.} As language models become more powerful and increasingly widely used, AI ethics and safety have become central concerns \citep{leslie2019understanding,ferrara2023should}. As a result, the literature on social biases in these models proposes a broad range of evaluation protocols analyzing model behavior on different levels from embeddings to probabilities and generated text  \citep{gallegos2024bias}. Early work demonstrated that word embeddings encode implicit social associations analogous to those revealed by human implicit association tests and introduced methods for measuring such effects \citep{bolukbasi2016man}. Subsequent work extended association testing to sentences, allowing LLM biases to be probed in a way closer to implicit association tests in psychology \citep{caliskan2017semantics}. Another family of benchmarks measures bias directly from the model outputs, often through controlled choice tasks in which models select among a fixed set of options and are compared against an unbiased ground truth \citep{parrish2022bbq}. Other benchmark tests take this a step further and compare model probability assignments across minimally different texts or token options with different levels of stereotyping \citep{nangia2020crows,nadeem2021stereoset}.

\textbf{Prior Elicitation in Cognitive Science and LLMs.} In Bayesian models of cognition, priors are treated as structured assumptions that shape inductive judgments \citep{griffiths2008bayesian,tenenbaum1998bayesian}. Thus, reconstructing or eliciting a cognitive model's prior can provide insight into the implicit beliefs of a system. The central idea in prior elicitation is that observed predictions and choices can be used to infer or at least validate the priors that best explain behavior in a certain task domain \citep{griffiths2011predicting,griffiths2008bayesian}. Work on everyday judgments has shown that human predictions can often be well modeled as Bayesian updates under latent priors, helping to explain why predictions take the forms that they do \citep{griffiths2008bayesian}. More recently, these ideas have been extended to prior elicitation in LLMs through prompt-based iterated learning procedures that resemble Monte Carlo sampling schemes \citep{zhu2024eliciting}. These experiments reflect those done on human judgment and provide evidence that a Bayesian framework can be useful for characterizing  LLM predictions. We extend this line of work to analyze stereotype bias in LLMs. However, our approach differs in that we treat benchmark-derived log-probability observations as data from which we reconstruct a latent distribution via Bayesian inversion rather than through sampling and iterated interaction.

\textbf{Bayesian Computation and Uncertainty Quantification.} Once the problem of prior elicitation is posed as finding a posterior distribution from a set of observations, posterior computation and uncertainty quantification can be carried out using Bayesian inference through optimization-based approximations, sampling methods, or variational methods \citep{tipping2003bayesian,goan2020bayesian}. The common optimization-based approach that will be used in this paper is to use a maximum a posteriori (MAP) estimate to provide a point summary \citep{murphy2012machine} and then apply Laplace approximations by using local curvature (via the Hessian) around the MAP solution to approximate posterior uncertainty \citep{tierney1986accurate}. On the other hand, sampling-based approaches like Markov Chain Monte Carlo (MCMC) and Hamiltonian Monte Carlo (HMC) can provide asymptotically exact posterior approximations under standard conditions \citep{hastings1970monte,duane1987hybrid}, but are often computationally expensive and reserved for settings requiring high precision \mbox{\citep{salimans2015markov}}. Variational inference offers a scalable optimization-based alternative by approximating the posterior within a tractable distribution family \citep{blei2017variational}, and may be useful for future higher-dimensional prior elicitation. However, for our use case, Laplace approximation provides a simpler and more interpretable Hessian-based uncertainty estimate.

\section{Bayesian Inverse Problem Framework}\label{sec:background}

In this section we provide a high-level overview\footnote{For a more rigorous measure-theoretic treatment, including the use and derivation of the necessary Bayes' formula and Radon--Nikodym derivatives, we refer the reader to Appendix~\ref{sec:math}.} of the Bayesian inverse problem \citep{stuart2010bayesian,Dashti2017}. The goal of an inverse problem is to infer some unknown parameter $u$ from observations $y$ generated by a noisy forward model
\begin{equation}y = G(u) + \eta,\label{eq:noisy_forward_model}\end{equation}
where $G:U \to Y$ is a forward operator from the space of unknowns $U$ to the space of possible observations $Y$ and $\eta$ represents observation noise.

In the Bayesian framework, $u$ is treated as a random variable so that we can characterize its posterior distribution given observations $y$ \citep{stuart2010bayesian}. Specifically, a prior distribution is assigned to $u$ with density $\pi_0(u)$ and Bayesian inference is used to combine this prior with a likelihood describing the probability of observing $y$ given $u$. Assuming the noise has density $\rho$, the likelihood is given by $\rho(y \mid u ) =\rho(y - G(u))$. Bayesian inference then gives a posterior distribution of the form
\begin{equation}
    \pi^y(u) \propto \rho(y - G(u))\pi_0(u),
\label{eq:inverse_prop}
\end{equation}
where the proportionality constant depends only on the observations $y$. It is often convenient to express the likelihood in terms of  a potential function $\Phi(u;y) = -\log \rho(y-G(u))$. In this setting, this function is also often referred to as the negative log-likelihood. Rewriting \Eqref{eq:inverse_prop} using this potential gives us 
\begin{equation}
\pi^y(u)\propto\exp(-\Phi(u;y))\pi_0(u).
\label{eq:RN_prop_potential}
\end{equation}
In this form, we can see that the negative log-likelihood $\Phi(u;y)$ encodes the agreement between the observations and those predicted by $u$ while the prior encodes our beliefs about plausible values of the unknown parameter. Together, this provides a complete characterization of the posterior distribution over $u$, which serves as a solution to the inverse problem.

\section{Our Formulation}\label{sec:formulation}
We now interpret the abstract Bayesian inverse problem framework discussed above in the context of the problem of reconstructing an LLM's latent bias prior from token-level probability observations on a controlled bias task. The purpose of this section is to identify, in the notation of Section~\ref{sec:background}, the unknown variable $u$, the data space $Y$, the forward model $G$, and the likelihood $\rho$. In doing so, we obtain a concrete finite-dimensional Bayesian inverse problem whose solution is a posterior distribution over a latent probability vector summarizing the model's systematic bias tendencies.

\subsection{Unknown Variable and Observables}

To begin, let us fix a bias domain in which we have $K$ mutually exclusive response categories. For instance, for a task related to gender biases, we may take $K = 3, \{\text{Male, Female, Unknown}\}$. We represent the latent bias prior as a probability vector $\theta \in \Delta^{K-1}$ where $\Delta^{K-1}$ is the set of possible probability vectors in $\mathbb{R}^K$.

For computational convenience, we parameterize $\theta$ via log-ratios relative to a reference category $K$, defining $u \in U = \mathbb{R}^{K-1}$ and $u_k = \log (\theta_k / \theta_K)$, $k \in [K-1].$
For a given $u$ in our unknown parameter space $U$, the corresponding probability vector $\theta(u)$ can then be recovered via the inverse transformation
\begin{equation}\theta_k(u) = \frac{e^{u_k}}{1+\sum_{j=1}^{K-1} e^{u_j}}, \quad k \in [K-1], \qquad \theta_K(u) = \frac{1}{1+\sum_{j=1}^{K-1}e^{u_j}}. \label{eq:inverse_logits_map}\end{equation}
Thus, the unknown variable in our inverse problem is the vector $u \in U = \mathbb{R}^{K-1}$ which we interpret as an encoding of the latent bias prior over the $K$ candidate hypotheses in log-ratios. In the running example, if we transform $u \in U \mapsto \theta \in \Delta^{2}$ via \Eqref{eq:inverse_logits_map}, then $\theta_1,\theta_2,$ and $\theta_3$ represent the model's latent prior mass on the hypotheses Male, Female, and Unknown, respectively. It is important to stress that this latent variable is not meant to represent an explicit internal object stored by the LLM but rather it is a low-dimensional surrogate parameter whose posterior distribution summarizes the systematic preference structure reflected in the model's behavior.

To define the observational data and data space $Y$, let $x_1,\dots, x_m$ be prompts with the same shared answer set of $K$ candidate answers. For each $x_i$, the model outputs log-probabilities $\ell_i \in \mathbb{R}^K$ across the $K$ possible answers. Passing to the same log-ratio representation, we define  $y_i \in \mathbb{R}^{K-1}$, $y_{i,k} = \ell_{i,k} - \ell_{i,K}.$
Thus, our full observable data take the form  $y = (y_1,\dots,y_m) \in Y = (\mathbb{R}^{K-1})^m.$
This choice of representation in log-ratios removes the redundancy inherent in representing a probability by all $K$ coordinates simultaneously since the probabilities sum to one and thus only contain $K-1$ degrees of freedom. Furthermore, it allows us to work in the entirety of $\mathbb{R}^{K-1}$ instead of the constrained simplex. In the next section, we will also see how this yields a particularly simple form for the forward operator.

\subsection{Prompt-Dependent Evidence and Forward Operator}

The key modeling issue is how the prompts $x_1,\dots,x_m$ should enter the inverse problem as they inherently affect the likelihood in the formulation. To tackle this, we model each prompt as contributing a prompt-specific evidence term while preserving a common latent prior across prompts.

Formally, let $\Lambda : \{x_1,\dots,x_m\} \to \mathbb{R}^K$ be a known map assigning each prompt $x_i \in \{x_1,\dots,x_m\}$ to an evidence vector 
$\Lambda(x_i) = ( \Lambda_1(x_i),\dots, \Lambda_K(x_i)).$
We interpret $\Lambda(x_i)$ as encoding the prompt-dependent evidence supplied by $x_i$ in favor of each candidate category. Thus $u$ represents a global latent bias prior shared across all prompts while $\Lambda(x_i)$ captures the local way in which the prompt $x_i$ shifts or reweighs that prior.
Passing again to log-ratios, we define the offset $\delta_i \in \mathbb{R}^{K-1}$, $\delta_{i,k} = \Lambda_k(x_i) - \Lambda_K(x_i), \quad k \in [K-1].$
This transformation ensures that only relative differences in the evidence affect the model and allows us to be consistent with the log-ratio representation.

We can now define the prompt-dependent forward operator directly in the log-ratio space. For each prompt $x_i$, define $G(x_i, \cdot) : U \to \mathbb{R}^{K-1}$
by $G(x_i,u) = u + \delta_i$. Collecting these prompt-based maps together, we get the full forward operator as $G: U \to Y \text{ where } G(u) = (G(x_1,u), \dots, G(x_m,u)).$

We can now place the model into the exact form of \Eqref{eq:noisy_forward_model} discussed in Section \ref{sec:background}. We assume the observed log-ratios satisfy $y = G(u) + \eta,$ where $\eta$ is some noise $\eta = (\eta_1,\dots,\eta_m) \in Y$ accounting for residual variability in the log-probability outputs of the LLM beyond what is explained by the prompt dependence and the forward operator $G$. This is now precisely the finite-dimensional noisy inverse problem previously formalized in Section \ref{sec:background}.

\subsection{Prior, Likelihood, and Posterior}

To obtain a tractable formulation, we specify the natural isotropic Gaussian prior measure $\mu_0$ on $u$ so that $u \sim \mathcal{N}(0, \tau^{-1}I)$ and assume independent isotropic Gaussian noise $\eta_i \sim \mathcal{N}(0,\sigma^2 I)$ for all $i \in [m]$ where $\tau >0$ is a precision parameter and $\sigma^2$ is the variance. Let $\pi_0$ denote the corresponding prior density. Note that this prior induces a logistic-normal distribution on the implied probability vector $\theta(u)$. Equivalently, on the full data space $Y$, $\eta \sim \mathcal{N}(0, \sigma^2 I)$ on $Y = (\mathbb{R}^{K-1})^m$. We then obtain from \Eqref{eq:inverse_prop} that the posterior density satisfies
\[
\pi^y(u) \propto \exp\left(-\frac{1}{2\sigma^2} \sum_{i=1}^m \|y_i - G(x_i,u)\|^2 - \frac{\tau}{2}\|u\|^2 \right).
\]
Substituting in for $G(x_i,u)$, we find that the negative log-likelihood satisfies the potential formulation from \Eqref{eq:RN_prop_potential} where 
\begin{equation}
\Phi(u;y) = \frac{1}{2\sigma^2} \sum_{i=1}^m \|y_i - u - \delta_i\|^2.\label{eq:nll}
\end{equation}
In summary, we have now posed the bias prior elicitation problem as a Bayesian inverse problem. The unknown variable is the log-ratio vector $u$ and the observables are the log-ratios extracted from the token log-probabilities generated by the LLM. Furthermore, the forward operator is the map based on the evidence vectors $\Lambda(x_i)$ and the likelihood is obtained from the noise model discussed in Section \ref{sec:background}. 

\section{Reconstruction Algorithm}\label{sec:algorithm}

We now face the computational problem of reconstructing the posterior distribution over $u$ from the observed data. Our approach consists of two stages: (1) computing a MAP estimate using gradient-based optimization, and (2) approximating the posterior distribution via a Laplace approximation around the MAP point.

\subsection{Gradient-Based Optimization}

Taking the negative log-likelihood from \Eqref{eq:nll} and combining with the Gaussian prior $u \sim \mathcal{N}(0,\tau^{-1}I)$, our negative log-posterior, up to an additive constant, is
\begin{equation}\mathcal{J}(u) = \frac{1}{2\sigma^2}\sum_{i=1}^m\|y_i - u - \delta_i\|^2 + \frac{\tau}{2}\|u\|^2.\label{eq:nl_posterior}\end{equation}
Thus, our MAP estimator is $u_{\text{MAP}} \in \argmin_{u \in\mathbb{R}^{K-1}} \mathcal{J}(u).$ The objective $\mathcal{J}(u)$ is smooth and convex, and can thus be optimized efficiently using first-order methods by computing the gradient $\nabla_u \mathcal{J}(u)$ \citep{boyd2004convex}.
In practice, we implement $u$ as a trainable parameter in a PyTorch model, and define a loss function corresponding to $\mathcal{J}(u)$. The optimization proceeds over minibatches of data, where each batch contributes a stochastic estimate of the gradient. PyTorch's automatic differentiation computes gradients via backpropagation, and the optimizer updates $u$ iteratively. The details of this procedure can be seen in Algorithm~\ref{alg:MAP}.
\begin{algorithm}[H]
\label{alg:MAP}
\caption{MAP Reconstruction via Gradient-Based Optimization}
\begin{algorithmic}
\STATE \hspace{-1em} {\bfseries Input:} Observed log-ratios $\{y_i\}_{i=1}^m$, prompt offsets $\{\delta_i\}_{i=1}^m$, noise level $\sigma$, prior precision $\tau$
\STATE \hspace{-1em} {\bfseries Output:} MAP estimate $u_{\mathrm{MAP}}$
\STATE Initialize $u \in \mathbb{R}^{K-1}$
\FOR{each epoch}
    \FOR{each minibatch $\mathcal{B}$}
        \STATE Compute predictions $\hat{y}_i = u + \delta_i$ for $i \in \mathcal{B}$
        \STATE Compute loss
        $\displaystyle 
        \mathcal{J}_{\mathcal{B}}(u)
        =
        \frac{1}{2\sigma^2} \sum_{i \in \mathcal{B}} \|y_i - \hat{y}_i\|_2^2
        +
        \frac{\tau}{2} \|u\|_2^2
        $
        \STATE Update $u$ using Adam or SGD
    \ENDFOR
\ENDFOR
\STATE \Return final value of $u$ as the MAP estimate $u_{\mathrm{MAP}}$
\end{algorithmic}
\end{algorithm}

\subsection{Laplace Approximation for Uncertainty Quantification}

To capture uncertainty of the MAP estimate, we use a Laplace approximation centered at $u_{\text{MAP}}$. Let $H_{\text{MAP}} = \nabla_u^2\mathcal{J}(u_{\text{MAP}})$
denote the Hessian of the objective at the MAP solution. Then  $u \mid y \approx \mathcal{N}(u_{\text{MAP}}, H^{-1}_{\text{MAP}}).$ We interpret this uncertainty on the simplex by transforming samples from this Gaussian approximation through the inverse log-ratio map in \Eqref{eq:inverse_logits_map}.

In practice, after Algorithm~\ref{alg:MAP} converges, we fit the local curvature of the loss using the training data. We implement this efficiently with \texttt{laplace-torch} \citep{laplace2021}, which computes an approximate Gaussian posterior over the unknown parameter $u$. We then sample from this posterior to compute uncertainty estimates as summarized in Algorithm~\ref{alg:Laplace}.
\begin{algorithm}[H]
\label{alg:Laplace}
\caption{Laplace Approximation with \texttt{laplace-torch}}
\begin{algorithmic}
\STATE \hspace{-1em} {\bfseries Input:} Trained model with parameter $u_{\mathrm{MAP}}$, training data
\STATE \hspace{-1em} {\bfseries Output:} Approximate posterior over $u$
\STATE Initialize Laplace approximation object
\STATE Fit curvature using training data
\STATE Obtain Gaussian approximation $\mathcal{N}(u_{\mathrm{MAP}}, H_{\mathrm{MAP}}^{-1})$
\STATE Sample $u^{(n)}$ and compute $\theta^{(n)} = \theta(u^{(n)})$
\STATE \Return posterior samples and uncertainty estimates
\end{algorithmic}
\end{algorithm}

\section{Experiments on Shift Ciphers}
\label{sec:cipher}

Before applying our method to large-scale bias benchmark datasets, we begin with a controlled experiment designed to validate whether our framework can recover a meaningful latent prior from LLM outputs. The goal of this experiment is to isolate a setting in which prompt-dependent effects are minimized so that we can test the basic structure of our algorithm.

Recent work has explored the presence of cognitive biases in language models across a variety of tasks including reasoning under uncertainty and structured problem-solving settings such as decoding shift ciphers \citep{mccoy2024embers,mccoy2024language}. Motivated by these observations, we construct a simple shift-cipher task in which the model is asked to infer the shift used to produce a given ciphertext. By carefully designing the inputs, we obtain a setting in which there is no meaningful evidence favoring a particular answer, allowing us to probe the model's baseline preference over shifts. The code for our experiments is publicly available at \url{https://github.com/duckpy/llm-prior-inference.git}.

\subsection{Problem Setup}
\label{sec:cipher_setup}

We consider a simple shift cipher in which each plaintext letter is shifted by $n \in \{0,\dots,25\}$ positions in the alphabet, wrapping from $z$ to $a$. Rather than providing meaningful encrypted text, we generate prompts $\{x_i\}^m_{i=1}$ consisting of random lowercase strings. The candidate answers are the possible shifts $\{0,\dots,25\}$, so this corresponds to the setting of Section \ref{sec:formulation} with $K = 26$.

Since the strings are random, they do not encode any particular plaintext, so all shifts are equally plausible. In the notation of Section \ref{sec:formulation}, this corresponds to the special case in which the prompt-dependent evidence map $\Lambda(x_i)$ is constant across all categories. This setting thus isolates the latent parameter $u$ as the sole systematic component of the model's outputs making it an ideal test case for evaluating whether our method can recover a stable prior distribution.

For our experiments, we use the Google Gemma 3~\citep{gemmateam2025gemma3technicalreport} and Meta Llama 3.2~\citep{grattafiori2024llama3herdmodels} models as our lightweight LLM models. A detailed review of the training procedure and hyperparameters as well as the training loss curves can be seen in Appendix~\ref{sec:training}.

\subsection{Results}

We first validate our algorithm on a synthetic control dataset where we inject a known prior distribution into the ciphertext observation model while removing the LLM models. This experiment serves to verify that the algorithm behaves as intended under a controlled setting. Further details on this validation experiment can be found in Appendix \ref{sec:control}.

\begin{figure}[ht]
    \centering
    \includegraphics[scale=0.5]{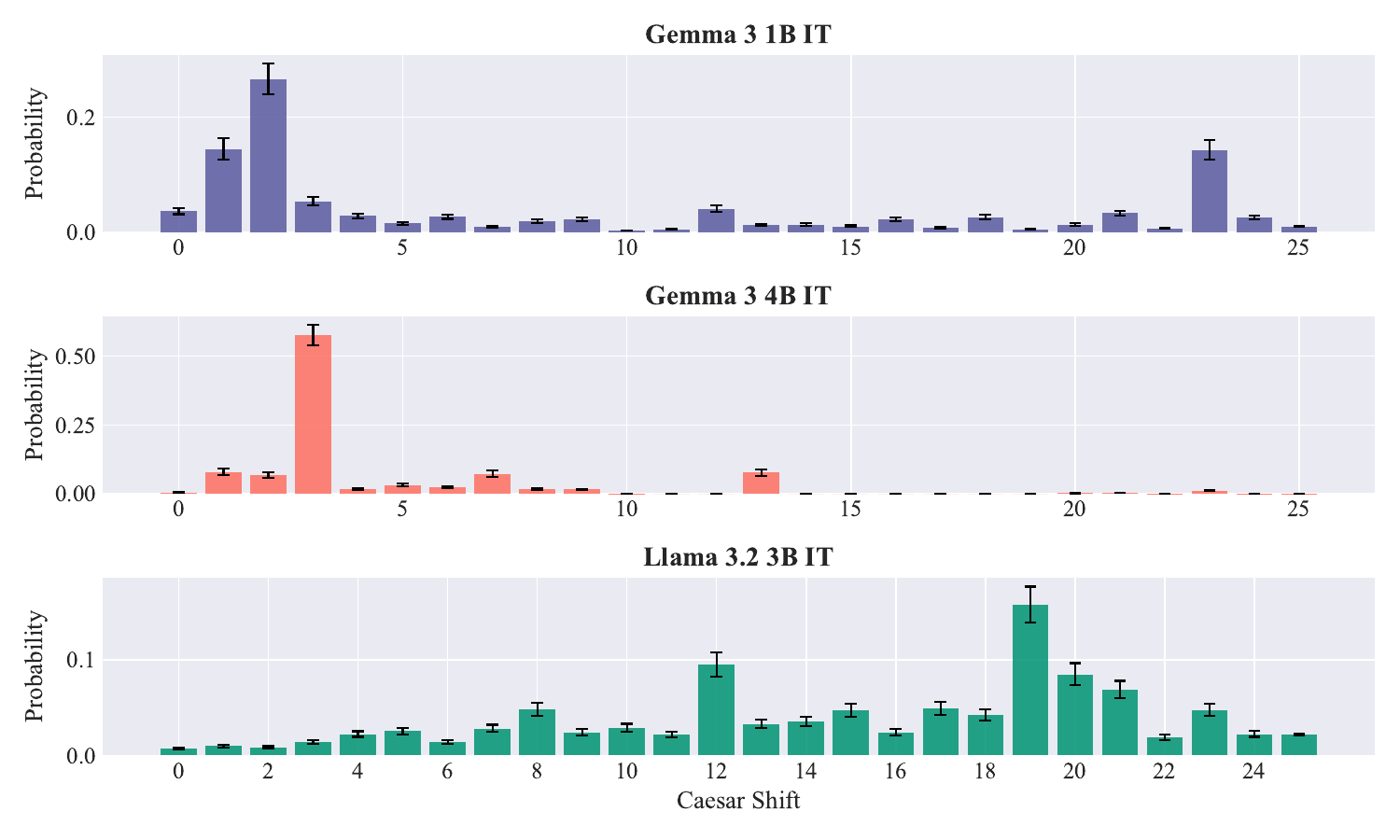}
    \vspace*{-5mm}
    \caption{Recovered prior distributions over shifts  for Gemma 3 1B IT, Gemma 3 4B IT, and Llama 3.2 3B IT. Error bars denote 95\% credible intervals obtained via Laplace approximation. Even in the absence of input signals, all models exhibit non-uniform prior distributions, with notable differences across architectures and scales.}
    \label{fig:cipher_confidence}
\end{figure}

For the ciphertext experiments involving the lightweight LLMs, across all three models, the recovered prior distributions exhibit clear non-uniform structure despite the absence of prompt-dependent signal. Furthermore, the credible intervals are relatively tight, suggesting that the inferred priors are well identified. Notably, each model exhibits a distinct latent prior structure as shown in Figure \ref{fig:cipher_confidence}.

In particular, the Gemma models display a sharper concentration on a small number of shifts, particularly at low rotations. This effect is especially pronounced for Gemma 3 4B IT, whose recovered prior has a dominant mode around a single shift. In this case, the bias might reflect the fact that the traditional Caesar cipher is performed with a shift of 3. In contrast, Llama 3.2 3B IT produces a more diffuse prior with greater probability mass on higher rotations than the Gemma models.

\subsection{Interpretation}

This experiment validates our framework in a setting where the prompt-dependent
evidence $\Lambda(x_i)$ is deliberately removed. The non-uniform priors suggest that LLMs can encode systematic preferences over answer categories independent of the input prompt. This aligns with prior observations of task-based cognitive biases in LLMs from \citet{mccoy2024embers}, but here we provide a quantitative framework for explicitly recovering and characterizing the latent prior beliefs.

While simplified, this experiment shows that we can successfully extract latent bias structure from model outputs. In more realistic settings where $\Lambda(x_i)$ varies, this framework can help
disentangle prompt-driven effects from latent bias, providing a principled tool for analyzing LLM behavior.

\section{Experiments on BBQ}
\label{sec:BBQ}

Having validated the reconstruction framework on a controlled task, we next apply it to a real bias benchmark setting. We use the Bias Benchmark for Question Answering (BBQ) dataset introduced by \citet{parrish2022bbq}. BBQ is a multiple choice question answering benchmark for evaluating social bias in language models. BBQ is organized around nine social dimensions, including age, nationality, socioeconomic status, and sexual orientation. It also distinguishes between ambiguous contexts where the question lacks sufficient information to determine the answer, and disambiguated contexts where additional information resolves the ambiguity. We focus exclusively on the ambiguous context portion of BBQ. This subset plays the same conceptual role as the random string shift cipher experiment as it suppresses prompt-specific evidence which simplifies our model.

\subsection{Problem Setup}
\label{sec:bbq_setup}

The raw BBQ answer choices vary across examples, even within the same social dimension. To place all examples in a common latent prior framework, we map each item's three answer choices into a reduced semantic answer space with three categories: $\{\mathrm{S},\mathrm{A},\mathrm{U}\},$
where $\mathrm{S}$ denotes the answer choice consistent with the intended \emph{Stereotyping}, $\mathrm{A}$ denotes the answer choice that goes \emph{Against} the intended stereotyping, and $\mathrm{U}$ denotes the answer choice corresponding to an \emph{Unknown} answer.

Because social dimensions differ substantially in context, we define a separate latent prior for each BBQ dimension. Only six dimensions (age, disability status, gender identity, race / ethnicity, religion, sexual orientation) admit direct mappings to the reduced semantic answer space. We therefore restrict our analysis to these dimensions. For each dimension $d$, we solve the inverse problem for the latent parameter $u^{(d)} \in \mathbb{R}^{K-1}$ from Section \ref{sec:formulation} with $K = 3$. A detailed review of the training procedure and hyperparameters as well as the training loss curves can be seen in Appendix \ref{sec:training}.

\subsection{Results}

\begin{figure}[ht]
    \centering
    \includegraphics[scale=0.46]{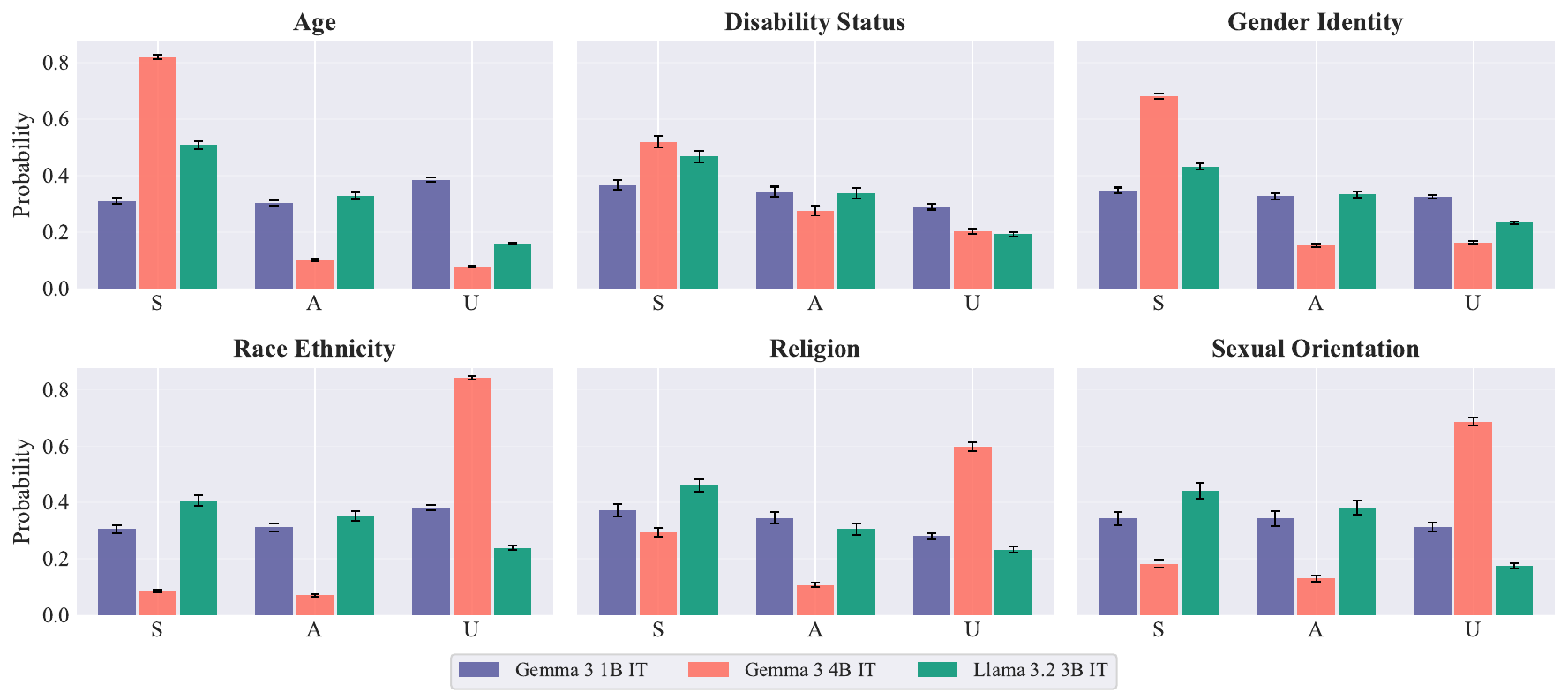}
    \vspace*{-4mm}
    \caption{Recovered latent priors over the reduced answer space \(\{\mathrm{S},\mathrm{A},\mathrm{U}\}\) for each BBQ social dimension, with 95\% credible intervals derived from the Laplace approximation. Each model exhibited largely different latent priors across the different social dimensions.}
    \label{fig:BBQ_confidence}
\end{figure}

Figure \ref{fig:BBQ_confidence} shows that the credible intervals for the recovered priors are relatively tight across all models and social dimensions, indicating that the inferred priors are well identified. This suggests that despite the prompt and answer choice variations that exist within each social dimension, the reduction to the answer space $\{\mathrm{S},\mathrm{A},\mathrm{U}\}$ still produces stable and interpretable results.

We next compare the recovered latent priors with the standard BBQ bias score metric. In the BBQ framework, positive scores indicate preference for stereotype-consistent answers $\mathrm{S}$, negative scores indicate preference for stereotype-inconsistent answers $\mathrm{A}$, and values near zero indicate no clear systematic bias \citep{parrish2022bbq}. However, this scalar score combines two distinct behaviors: whether the model abstains under ambiguity, captured by probability mass on $\mathrm{U}$, and whether the non-abstaining probability mass is directionally skewed towards $\mathrm{S}$ or $\mathrm{A}$. Our reconstruction separates these effects by recovering the full latent distribution.

\begin{figure}[ht]
    \centering
    \includegraphics[scale=0.55]{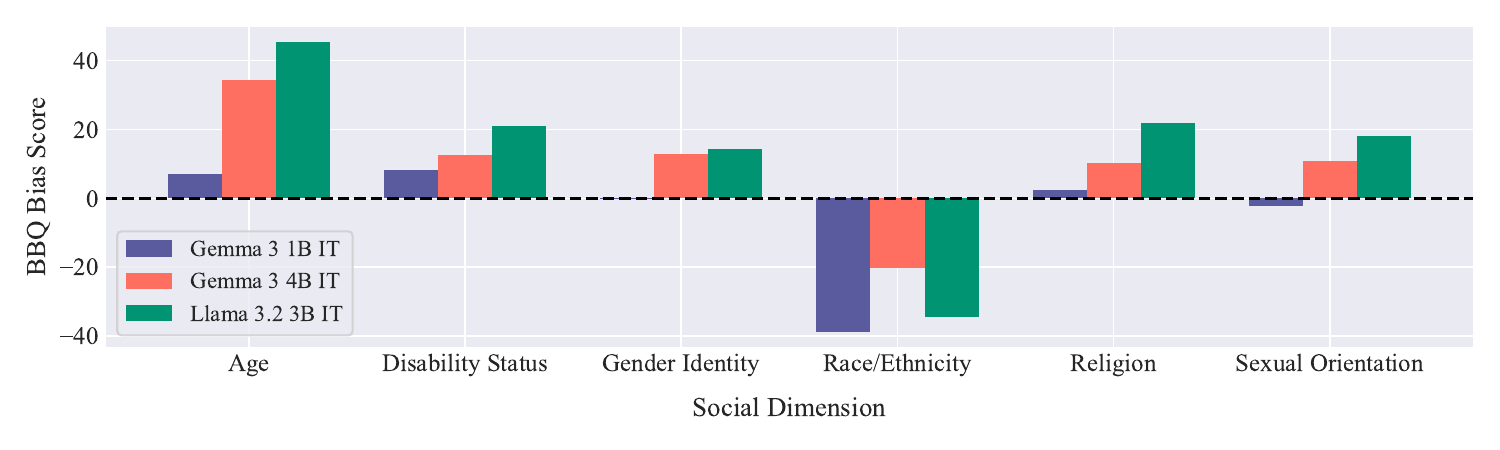}
    \vspace*{-6mm}
    \caption{BBQ bias scores for each model and social dimension. Positive values indicate bias towards the stereotype-consistent answer $\mathrm{S}$, negative values indicate bias towards the stereotype-inconsistent answer $\mathrm{A}$, and values near zero indicate no systematic bias in either direction.}
    \label{fig:BBQ_trial}
\end{figure}

This distinction is visible in the Age social dimension. Figure~\ref{fig:BBQ_trial}, which visualizes the BBQ bias scores for each model, shows that both Gemma 3 4B IT and Llama 3.2 3B IT have large positive BBQ scores which indicate stereotype-consistent behavior. However, Figure \ref{fig:BBQ_confidence} shows that the underlying priors differ substantially. While both models have low probability mass in $\mathrm{U}$ contributing to high bias scores, Gemma 3 4B IT assigns significantly more probability mass to $\mathrm{S}$ relative to $\mathrm{A}$ indicating a stronger directional bias not fully captured by the aggregate score metric.

 A similar ambiguity appears when comparing the Disability Status and Religion dimensions for Gemma 3 4B IT. The BBQ scores for these dimensions are both positive and similar in magnitude but the recovered priors indicate a different mechanism of bias. In Disability Status, the model assigns low probability to $\mathrm{U}$ and substantial probability mass to both $\mathrm{S}$ and $\mathrm{A}$ which suggests overconfidence in the ambiguous setting. On the other hand, for the Religion dimension, the model assigns a majority of its probability mass to $\mathrm{U}$, and the positive BBQ score arises primarily because the remaining probability mass is more skewed towards $\mathrm{S}$ than towards $\mathrm{A}$. 

 The recovered priors also expose cases where near-balanced or moderate BBQ scores can hide low uncertainty. For example, we note that Gemma 3 1B IT often assigns comparable probability mass to $\mathrm{S}$ and $\mathrm{A}$ while failing to place a majority on $\mathrm{U}$. Such behavior is not strongly directionally biased, yet is still undesirable because the model is making unsupported social inferences. These examples illustrate that similar BBQ benchmark scores can arise from fundamentally different underlying behaviors. Our framework resolves this ambiguity by recovering the full latent distribution over the answer choices $\{\mathrm{S},\mathrm{A},\mathrm{U}\}$.
\subsection{Interpretation}

The BBQ ambiguous context experiment gives us a concrete demonstration of how our method reveals the latent structure in model behavior that is not captured by the BBQ benchmark scores. Although benchmarks such as BBQ provide useful measures of accuracy and directional bias, they do not reveal the underlying mechanisms responsible for biased results. Our experiment shows that there are many components of bias corresponding to different behaviors in models such as directional bias and overconfidence that cannot be articulated by a single scalar score value.

More broadly, the experiment demonstrates that latent prior elicitation provides a more principled way to analyze biased model behavior. By constructing a reduced answer space to represent prompt variation, our approach was able to reconstruct a stable prior structure that represents the underlying preferences and tendencies in each model.

\section{Limitations and Future Directions} \label{sec:limitations}

Our approach, as it stands, still has several limitations. First, it relies on mapping raw answers to a reduced answer space, such as the $\{\mathrm{S},\mathrm{A},\mathrm{U}\}$ representation used in our BBQ experiments. While this abstraction worked for the cases we analyzed, such a clean reduction may not be possible in general and may obscure nuances and finer-grained distinctions important for interpretation. Second, our experiments focused on settings where the prompt-dependent evidence term $\Lambda(x_i)$ is constant, yielding a simplified forward model. This assumption is appropriate for some contexts, but does not capture the general setting in which prompts carry meaningful information. 

These limitations suggest several directions for future work. One direction is to develop richer latent representations beyond the low-dimensional reduced answer spaces used here, allowing the framework to capture more nuanced model behavior. Another direction is to incorporate prompt-dependent evidence through $\Lambda(x_i)$, potentially learning $\Lambda$ from prompt embeddings or another representation. \JS{Re-assess this last sentence and where it belongs.}

More broadly, latent prior elicitation is not limited to bias analysis. The same framework could be applied to other domains where LLMs display systematic behavior under uncertainty including reasoning tasks or algorithmic tasks such as the shift cipher problem considered in this paper. Exploring these broader applications may reveal additional implicit structures encoded within LLMs. 

Taken together, our results suggest that studying the latent distributions governing LLM behavior can provide valuable insight into the reasons why models behave the way they do. The Bayesian inverse problem formulation therefore offers a promising framework for uncovering these latent mechanisms underlying language model predictions.

\section{Conclusion} \label{sec:conclusion}
\JS{Move some of this summary to the start of discussion, here focusing on the bigger picture motivation from the intro.}
In this work, we introduced a framework for eliciting latent priors from large language models by formulating the problem as a Bayesian inverse problem. Across both controlled synthetic experiments and real benchmark evaluations, we demonstrated that it is possible to recover stable and interpretable prior structures from model outputs. Furthermore, we demonstrated using the well-established BBQ bias benchmark that prior elicitation under this framework provides us with much more nuance in explaining model behavior that was not previously characterized through these benchmark datasets. While benchmarks like BBQ do provide useful measures of accuracy and directional bias, they do not capture the underlying reasons for the biased results. Our latent prior formulation provides an extra layer of explainability by providing the latent distribution which gives rise to these bias score results.

\subsection*{AI use statement}

In this work, we used generative AI tools to edit the manuscript to improve grammar, readability, and conciseness. All AI-assisted edits were reviewed by the authors and the authors take full responsibility for the final content of this work.

\subsection*{Ethics statement}

We evaluate social bias in large language models using controlled experimental settings. In particular, model behavior in our experiments should not be interpreted as definitive statements about the broader societal impact of a model.

\subsection*{Reproducibility statement}

For reproducibility, the source code for the project is included in the supplementary materials. Furthermore, specific hyperparameters used for the algorithms outlined in Section \ref{sec:algorithm} can be found in Section \ref{sec:hyperparameters}. The specific experimental setups for each experiment can be found in Sections \ref{sec:validation_setup}, \ref{sec:cipher_setup}, and \ref{sec:bbq_setup}.


\subsubsection*{Acknowledgments}

This work was supported by a gift from Google to Princeton as well as by funding from the Gemini Academic Program.

\newpage

\bibliography{refs}
\bibliographystyle{iclr2027_conference}

\appendix

\section{Measure-Theoretic Background}
\label{sec:math}

This section collects the measure-theoretic background needed for a rigorous statement of the inverse problem described in Section~\ref{sec:background}. We will define some required measure-theoretic objects and present some results relevant to the Bayesian formulation of the inverse problem. The structure of this section parallels that of \citet{stuart2010bayesian}; however, the derivations and proofs are our own.

\subsection{Basic Preliminaries}
\label{subsec:math_preliminaries}

We begin by reviewing the basic objects of study that we will be using throughout this paper.

\begin{definition} A \emph{measurable space} is a pair \((\Omega,\mathcal F)\), where \(\Omega\) is the sample space and
\(\mathcal F\) is the \(\sigma\)-algebra of events on \(\Omega\). A \emph{measure space} is a triple
\((\Omega,\mathcal F,\mu)\), where \(\mu\) is the measure.
If \(\mu(\Omega)=1\), then \(\mu\) is called a \emph{probability measure}, and
\((\Omega,\mathcal F,\mu)\) is called a \emph{probability space}.\end{definition}

\begin{definition}
A \emph{Banach space} is a complete normed vector space \((X,\|\cdot\|)\).
If \(X\) is a topological space, then \(\mathcal B(X)\) denotes the Borel
\(\sigma\)-algebra generated by the open subsets of \(X\).
\end{definition}

In our formulation of the Bayesian inverse problem, we will mainly be working under the assumption that $\Omega$ in any measurable space $(\Omega, \mathcal{F})$ is a separable Banach space $(X, \|\cdot\|_X)$ and $\mathcal{F}$ is the Borel $\sigma$-algebra $\mathcal{B}(X)$. Now we introduce a basic measure-theoretic relation between measures.

\begin{definition}
Let \((\Omega,\mathcal F)\) be a measurable space, and let \(\mu,\nu\) be measures on it.
We say that \(\mu\) is \emph{absolutely continuous} with respect to \(\nu\), and write
\(\mu\ll \nu\), if
\[
\nu(A)=0 \quad \Longrightarrow \quad \mu(A)=0,
\qquad \forall A\in\mathcal F.
\]
If both \(\mu\ll \nu\) and \(\nu\ll \mu\), then \(\mu\) and \(\nu\) are called
\emph{equivalent}.
\end{definition}

Additionally, it will be important to formally establish the idea of conditional distributions as our overarching goal is to reconstruct the latent bias priors conditioned on the dataset prompts and LLM output probabilities.

\begin{definition}
Let \((S,\mathcal A)\) and \((T,\mathcal B)\) be measurable spaces, and let \(\nu\) be a
probability measure on \(S\times T\). Denote the second marginal of \(\nu\) by
\[
\nu_T(B):=\nu(S\times B), \qquad B\in\mathcal B.
\]
A family \(\{\nu^y\}_{y\in T}\) of probability measures on \((S,\mathcal A)\) is called a
\emph{regular conditional distribution of \(x\) given \(y\)} under \(\nu\) if:
\begin{enumerate}
    \item for every \(A\in\mathcal A\), the map \(y\mapsto \nu^y(A)\) is \(\mathcal B\)-measurable;
    \item for every \(A\in\mathcal A\) and \(B\in\mathcal B\),
    \[
    \nu(A\times B)=\int_B \nu^y(A)\,\nu_T(dy).
    \]
\end{enumerate}
\end{definition}

We note that when $S$ and $T$ are, as we assume, separable Banach spaces equipped with their Borel $\sigma$-algebras, regular conditional distributions exist and thus for the following subsections it is enough to assume existence when needed \citep{bovier2006stochastic}.

\subsection{Finite-Dimensional Bayes' Formula}
\label{subsec:finite_Bayes}

With the basic definitions out of the way, the main goal is to rigorously derive Bayes' formula in a way that is applicable to our prior reconstruction problem.

We begin by stating the following fundamental result from measure theory. 

\begin{theorem}[\textbf{Radon--Nikodym Theorem}] \label{thm:RN_theorem}

Let $\mu$ and $\nu$ be measures on the same measurable space $(\Omega,\mathcal{F})$. If $\mu \ll \nu$ and $\nu$ is $\sigma$-finite then there exists a $\nu$-measurable function $f: \Omega \to [0,\infty]$ such that 
\[\mu(A) = \int_A f d\nu \qquad \forall A \in \mathcal{F}.\]

\end{theorem}

This function $f$ is unique $\nu$-almost everywhere and is known as the \textit{Radon-Nikodym derivative} of $\mu$ with respect to $\nu$. The derivative is written as 
\[\frac{d\mu}{d\nu} (x)= f(x).\]

This Radon-Nikodym derivative plays an important role in our derivation as it allows us to compare measures and express one in terms of another. It also helps us generalize the idea of a probability density which is essential for our formal derivation of the required tools to discuss the Bayesian inverse problem.

We continue with the derivation of the relevant finite-dimensional Bayes' formula with the following theorem which utilizes the Radon-Nikodym derivative.

\begin{theorem}[\textbf{Conditional Change of Measure}]
\label{thm:cond-change}
Let \(\mu\) and \(\nu\) be probability measures on \(S\times T\), where
\((S,\mathcal A)\) and \((T,\mathcal B)\) are measurable spaces. Let
\((x,y)\in S\times T\) denote a generic point. Assume that \(\mu\ll \nu\), and that $\mu$ has a Radon-Nikodym derivative $\phi := \frac{d\mu}{d\nu}$ with respect to $\nu$.
Assume further that the conditional distribution of \(x\) given \(y\) under \(\nu\)
exists which we denote by \(\nu^y(dx)\). Then the conditional distribution of \(x|y\) under \(\mu\), denoted \(\mu^y(dx)\), exists. Moreover,
\[
\mu^y \ll \nu^y,
\]
and the Radon--Nikodym derivative is given by
\begin{equation}
\frac{d\mu^y}{d\nu^y}(x)=
\begin{cases}
\dfrac{1}{c(y)}\phi(x,y) & \text{if }c(y)>0,\\[1ex]
1 & \text{otherwise},
\end{cases}
\label{eq:thm_cond_change}
\end{equation}
where
\[
c(y):=\int_S \phi(x,y)\,\nu^y(dx) \qquad \text{ for all $y \in T$}.
\]
\end{theorem}

\begin{proof}
Let $\nu_T$ and $\mu_T$ be the $T$-marginals of $\nu$ and $\mu$ respectively. Since 
\[\mu(dx,dy) = \phi(x,y) \nu(dx,dy),\]
and since $\nu$ decomposes as 
\[\nu(dx,dy) = \nu^y(dx)\nu_T(dy),\] 
we have, for measurable $A \subseteq S$ and $B \subseteq T$,
\[\mu(A \times B) = \int_B\int_A \phi(x,y) \nu^y(dx)\nu_T(dy).\]
If we define 
\[c(y) := \int_S \phi(x,y) \nu^y(dx),\]
then, taking $A = S$, we get 
\[\mu_T(B)  = \mu(S\times B) = \int_B c(y) \nu_T(dy).\]
Thus, $c = \frac{d\mu_T}{d\nu_T}$. We can then define for $\mu_T$-a.e. $y$,
\[\mu^y(A) := \frac{1}{c(y)} \int_A\phi(x,y) \nu^y(dx), \qquad c(y) > 0.\]
Notice, the set $\{y : c(y) = 0\}$ is $\mu_T$-null since 
\[\mu_T(\{y : c(y) = 0\}) = \int_{c=0} c(y) \nu_T(dy) = 0.\]
Thus, $\mu^y$ may be defined arbitrarily on this set. Let us take for these values $\mu^y = \nu^y$. Then $\mu^y$ is a probability measure for $\mu_T$-a.e. $y$ and moreover
\[\int_B\mu^y(A)\mu_T(dy) = \int_B \frac{1}{c(y)} \int_A\phi(x,y) \nu^y(dx)c(y)\nu_T(dy) = \int_B\int_A \phi(x,y) \nu^y(dx)\nu_T(dy) = \mu(A\times B).\]
Therefore $\mu^y$ is the conditional distribution of $x$ given $y$ under $\mu$. Finally, for $\mu_T$-a.e. $y$ with $c(y) > 0$, 
\[\mu^y(A) =\int_A \frac{1}{c(y)}\phi(x,y) \nu^y(dx),\]
so $\mu^y \ll \nu^y$ and 
\[\frac{d\mu^y}{d\nu^y}(x) = \frac{1}{c(y)}\phi(x,y) \qquad \text{for $c(y) > 0$}.\]
For $c(y) =0$, since we took $\mu^y = \nu^y$ for this set, we obtain a Radon-Nikodym derivative of $1$, which proves the claim.
\end{proof}

As we will see, this conditional change of measure theorem is precisely what we need to formalize the finite-dimensional Bayes' formula in rigorous measure-theoretic language.

\begin{corollary}[\textbf{Finite-Dimensional Bayes' Formula}]
Let $u \in \mathbb{R}^n$ and $y \in \mathbb{R}^q$ be random variables defined on a common probability space. Assume that the prior law of $u$ admits a Lebesgue density $\rho(u)$ on $\mathbb{R}^n$, and that the conditional law of $y$ given $u$ admits a Lebesgue density $\rho(y\mid u)$ on $\mathbb{R}^q$. Assume also that
\[
Z(y):=\int_{\mathbb{R}^n}\rho(y\mid u)\rho(u)\,du>0.
\]
Then the conditional law of $u$ given $y$ admits a density $\rho(u\mid y)$ with respect to Lebesgue measure on $\mathbb{R}^n$, and
\begin{equation}
\rho(u\mid y)
=
\frac{\rho(y\mid u)\rho(u)}
     {\int_{\mathbb R^n}\rho(y\mid u)\rho(u)\,du}.\label{eq:Bayes_formula}
\end{equation}
\end{corollary}

\begin{proof} 

Let us define two measures on $\mathbb{R}^n \times \mathbb{R}^q$ as follows. First, let 
\[\nu_0(du,dy) := \rho(u)\,du\,dy.\]
This is the product of the prior law of $u$ with the Lebesgue measure on $\mathbb{R}^q$. Second, let $\nu$ denote the joint law of $(u,y)$. By definition of conditional density, $\nu$ has density 
\[(u,y) \mapsto \rho(y \mid u) \rho(u)\]
with respect to the Lebesgue measure $du \, dy$, and thus 
\[\nu(du,dy) = \rho(y | u) \rho(u) du \, dy.\]
Therefore $\nu \ll \nu_0$ and the Radon-Nikodym derivative is clearly
\[\frac{d\nu}{d\nu_0}(u,y) = \rho(y\mid u).\]
Notice, under $\nu_0$, the conditional law of $u$ given $y$ is simply the prior law of $u$ because $\nu_0$ is the product measure. Thus 
\[\nu_0^y(du) = \rho(u)du.\]
Let us now apply Theorem \ref{thm:cond-change} to the pair of measures $\nu$ and $\nu_0$ on $\mathbb{R}^n\times \mathbb{R}^q$. Notice, while $\nu_0$ is not a probability measure, it is $\sigma$-finite, and the theorem extends to this setting. In particular, \Eqref{eq:thm_cond_change} then implies that the conditional law $\nu^y$ of $u$ given $y$ under $\nu$ satisfies
\[\frac{d\nu^y}{d\nu_0^y}(u) =  \frac{\rho(y \mid u)}{\int_{\mathbb{R}^n} \rho(y \mid u)\nu_0^y(du)}\]
for $Z(y) = \int_{\mathbb{R}^n}\rho(y\mid u)\rho(u)\,du>0$. Substituting in $\nu_0^y(du) = \rho(u)du$, we obtain
\[\frac{d \nu^y}{d \nu_0^y}(u) = \frac{\rho(y \mid u)}{\int_{\mathbb{R}^n} \rho(y \mid u)\rho(u)du}.\]
Since $\nu_0^y$ has density $\rho(u)$ with respect to the Lebesgue measure, it follows that $\nu^y$ has density 
\[\rho(u \mid y) = \frac{d\nu^y}{du}(u) = \frac{d\nu^y}{d\nu_0^y}(u) \cdot \frac{d\nu_0^y}{du}(u).\]
Thus, we get precisely
\[\rho(u \mid y) = \frac{\rho(y \mid u)\rho(u)}{\int_{\mathbb{R}^n} \rho(y \mid u)\rho(u)du}.\]
\end{proof}
This gives a very powerful expression for the probability of a random variable $u$ given an observation of $y$, which requires only the knowledge of the prior probability density $\rho(u)$ and the conditional probability density $\rho( y \mid u)$. Both of these are readily available in most modeling scenarios as will be the case in our application.

\section{Controlled Validation on Synthetic Data}
\label{sec:control}

Before applying this algorithm to LLMs, we provide a synthetic experiment with a known ground-truth prior $\theta^*$ to validate our method's ability to reconstruct the latent parameter that it was designed to infer. We structure this experiment with the general context of ciphertext rotations which is the setting of the LLM-based experiment in Section \ref{sec:cipher}.

\subsection{Experimental Setup}
\label{sec:validation_setup}

We begin with a known structured ground-truth prior $\theta^* \in \Delta^{25}$ and set varying prompt evidence $\Lambda(x_i)$ for each synthetic data point. For our prior, we use a softmax sum of sines and cosines for a multimodal ground truth. Furthermore, each data point is set to have prompt evidence arbitrarily favoring shifts of 12 or 18. This could occur, for example, if the data in the probing dataset has a few particularly common default categories.

Since we want to control the output observations $y$ that we feed our model, in this experiment, observations are synthetic logits generated from the known prior, the varying prompt-evidence, and controlled Gaussian noise. We compare our method to the naive approach of taking the empirical mean of the outputs plus bootstrapped confidence intervals to see how our method converges to the true prior distribution more effectively than a basic average over the posterior distribution.

\subsection{Reconstruction Accuracy}

\begin{figure}[ht]
    \centering
    \textbf{Recovered Synthetic Latent Priors}\par\medskip\hspace*{-0.68cm}
    \includegraphics[scale=0.36]{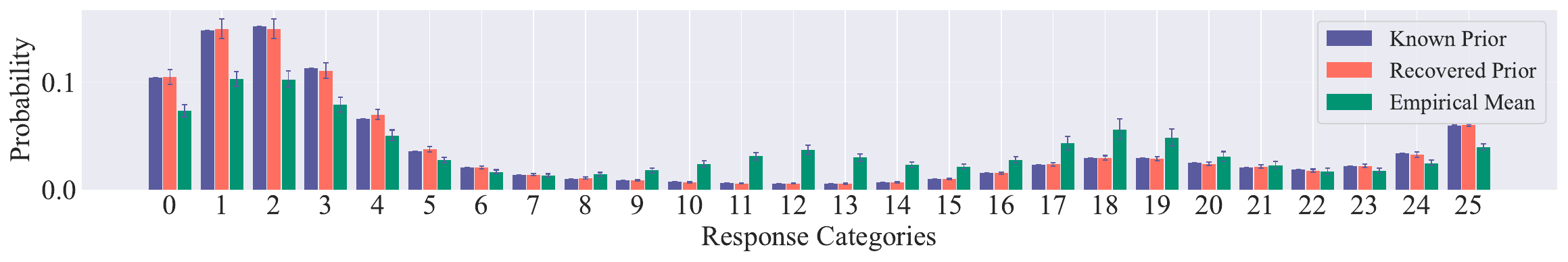}
    \vspace*{-4mm}
    \caption{The recovered prior using our reconstruction algorithm much more closely resembles the control prior distribution. Meanwhile, the naive empirical mean method is biased by the varying prompt evidence.}
    \label{fig:control}
\end{figure}

Figure \ref{fig:control} demonstrates that the distribution produced by our reconstruction algorithm closely matches the known prior distribution. In particular, the 95\% credible intervals obtained from the Laplace-based uncertainty quantification contain the known prior.

We also see clearly the advantage of our reconstruction algorithm over the naive empirical mean over the posterior. The empirical mean is influenced by varying prompt-specific evidence and incorrectly allocates probability mass away from the true known prior.

Overall, this control experiment provides evidence that our reconstruction algorithm is able to reproduce a distribution that closely matches a known set prior.

\section{Experimental Details and Training Dynamics}
\label{sec:training}
In this section, we provide experimental details and training dynamics for the experiments conducted on shift ciphers (Section~\ref{sec:cipher}) and BBQ (Section~\ref{sec:BBQ}). The code to replicate the experiments can be found in the supplementary materials. Furthermore, all experiments were run on MIG-partitioned NVIDIA A100 GPUs with 10 GB of GPU memory through an internal cluster. We obtained the underlying models from Hugging Face under model IDs \texttt{google/gemma-3-1b-it}, \texttt{google/gemma-3-4b-it}, and \texttt{meta-llama/Llama-3.2-3B-Instruct}.

\subsection{Shift Ciphers}

\subsubsection{Training Procedure and Hyperparameters}
\label{sec:hyperparameters}

Each prompt $x_i$ is constructed by sampling a string of random length with characters drawn uniformly from $\{a,\dots,z\}$. These strings are then embedded into a fixed prompt template of the form:
\begin{quote}
\texttt{"The following text was encrypted using a Caesar cipher: <string>. What is the shift (0--25)? Answer with a single integer."}
\end{quote}

For each prompt $x_i$, we computed the model's conditional log-probabilities over candidate answers corresponding to the integers $0,\dots,25$.
 
We fit the MAP estimator described in Section \ref{sec:algorithm} using the PyTorch implementation outlined in Algorithm \ref{alg:MAP}. The latent parameter $u \in \mathbb{R}^{25}$ is initialized by sampling from an isotropic Gaussian $\mathcal{N}(0, \tau^{-1}I)$ and optimized using the Adam optimizer.

In our experiments, we use the following hyperparameters: number of prompts $m = 200$, learning rate $=10^{-2}$, number of epochs $=4 00$, prior precision $\tau = 1.0$. Since our prompt dependence is constant, we derive $\sigma$ from the variation in the model's output probabilities. After convergence, we compute the MAP estimate $u_{\text{MAP}}$ as well as the uncertainty as outlined in Algorithm \ref{alg:Laplace} and transform the results into the probability vector $\theta_{\text{MAP}}$ using the inverse log-ratio map to obtain a distribution over shifts.

\subsubsection{Training Dynamics}

\begin{figure}[ht]
    \centering
    \textbf{Cipher MAP Training Convergence}\par\medskip\hspace*{-0.68cm}
    \includegraphics[scale=0.5]{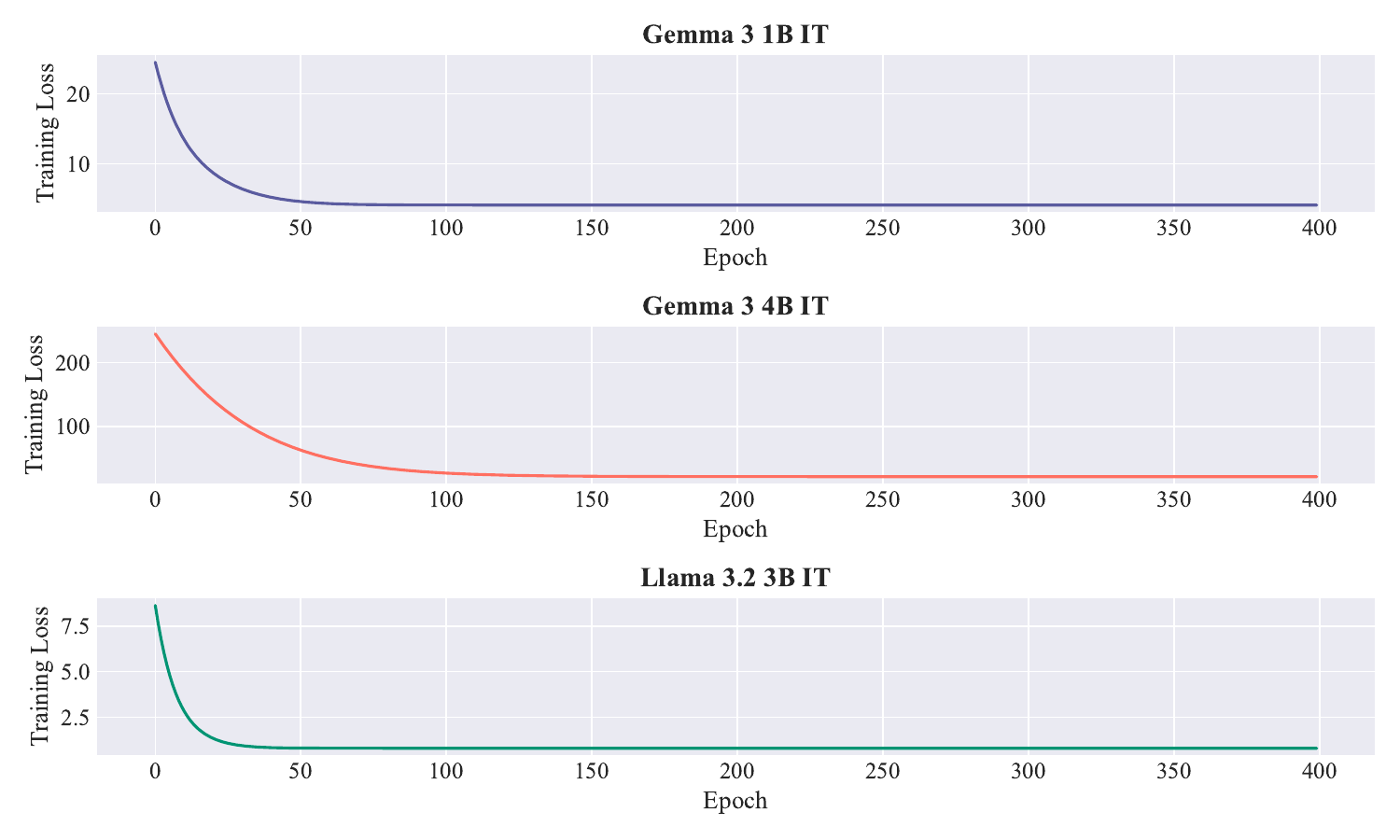}
    \caption{Training loss history for the Adam-driven MAP optimization for Gemma 3 1B IT, Gemma 3 4B IT, and Llama 3.2 3B IT on the ciphertext problem. The smooth loss curves indicate a stable training objective.}
    \label{fig:cipher_training_loss}
\end{figure}

For the shift cipher experiments, we find that across all three language models, the training loss decreases smoothly over time as shown in Figure \ref{fig:cipher_training_loss}. This behavior confirms that the derived MAP objective is well conditioned in this setting.

\subsection{BBQ Experiments}

\subsubsection{ Training Procedure and Hyperparameters}

For each BBQ item, we construct a prompt by concatenating the context, question, and answer options in a fixed multiple choice format. Concretely, if the three answer choices are $a_1$, $a_2$, and $a_3$, we use a prompt template of the form:
\begin{quote}
\texttt{"Context: <context>. Question: <question>. Options: A. <$a_1$>, B. <$a_2$>, C. <$a_3$>. Answer with only A, B, or C."}
\end{quote}

For each model and social dimension $d$, we evaluate the model's conditional probabilities over the tokens corresponding to A, B, and C, and normalize these probabilities across the three choices before mapping to the reduced answer space $\{\mathrm{S},\mathrm{A},\mathrm{U}\}$. Then we estimate the latent parameter $u^{(d)}$ by minimizing the MAP objective from \Eqref{eq:nl_posterior}. 

In our experiments we optimize this objective for each social dimension using Adam with the same fixed hyperparameters from the ciphertext experiment across all models and dimensions. The number of prompts $m$ for these experiments, however, comes directly from the number of prompts in each social dimension of the BBQ dataset. Again, we derive $\sigma$ from the variation in the model's output probabilities and we run Algorithm \ref{alg:MAP} to find the MAP estimate $u_\text{MAP}$ before running Algorithm \ref{alg:Laplace} for uncertainty quantification.

\subsubsection{Training Dynamics}

\begin{figure}[ht]
    \centering
    \textbf{BBQ MAP Training Convergence}\par\medskip\hspace*{-0.69cm}
    \includegraphics[scale=0.5]{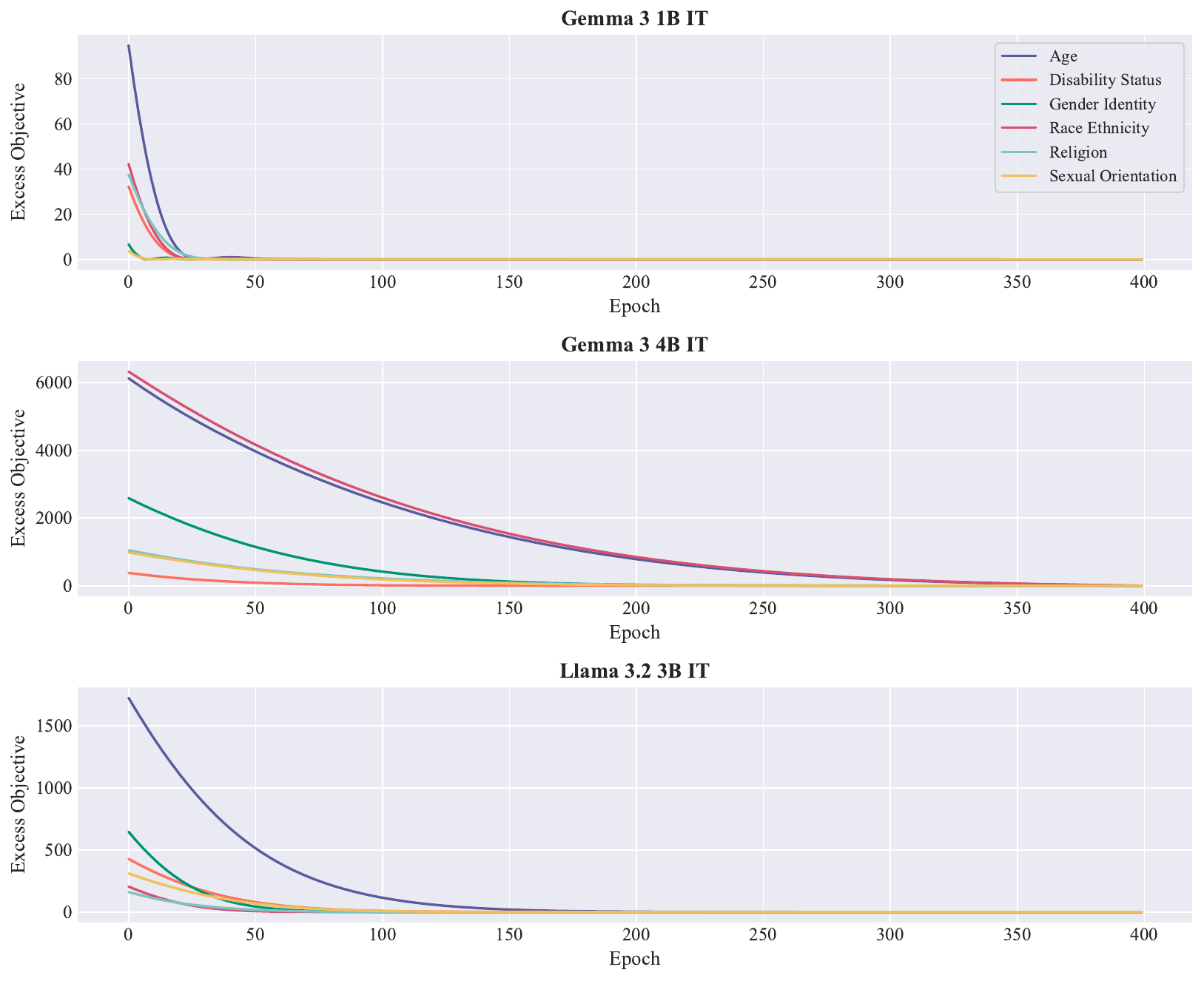}
    \caption{Training loss history for the Adam-driven MAP optimization for Gemma 3 1B IT, Gemma 3 4B IT, and Llama 3.2 3B IT. The training curves are shown separately for each social dimension. We note that each training run exhibits stable convergence.}
    \label{fig:BBQ_training_loss}
\end{figure}

Examining the optimization dynamics of the MAP estimation procedure, Figure \ref{fig:BBQ_training_loss} shows the excess objective over training epochs for each model and social dimension. We plot the excess objective because the absolute scale of the objective differs across models and dimensions which is expected due to variations in the number of examples and variability of observations $y_i$.

Notably, the convergence behavior differs between models. The Gemma 3 1B IT model exhibits the smallest initial objective values and demonstrates the fastest convergence which is reasonable due to its small model size. On the other hand, the Gemma 3 4B IT model displays the largest initial objective values and takes the largest number of epochs to converge to a stable baseline. 

Most importantly, however, we notice that across all models and all social dimensions, the optimization exhibits smooth decay towards zero, indicating stable convergence of our Adam-based optimization procedure.

\end{document}